\documentclass[journal]{IEEEtran}
\ifCLASSINFOpdf
   \usepackage[pdftex]{graphicx}
   \graphicspath{{../pdf/}{../jpeg/}}
   \DeclareGraphicsExtensions{.pdf,.jpeg,.png}
\else
   \usepackage[dvips]{graphicx}
   \graphicspath{{../eps/}}
   \DeclareGraphicsExtensions{.eps}
\fi
\usepackage{booktabs}
\usepackage{amsmath}
\usepackage{amsfonts}
\usepackage{cite}
\begin{document}
%
\title{Depth-Assisted ResiDualGAN for Cross-Domain Aerial Images Semantic Segmentation}
%
%
%

\author{\IEEEauthorblockN{Yang Zhao,
Peng Guo,
Han Gao, and 
Xiuwan Chen}
\thanks{Y. Zhao, P. Guo,  H. Gao, and X. Chen are with the Institute of Remote Sensing and Geographic Information System, Peking University, Beijing 100871, China. E-mail: zy\_@pku.edu.cn; peng\_guo@pku.edu.cn; hgao@pku.edu.cn; xwchen@pku.edu.cn.}
\thanks{Corresponding author:Han Gao(hgao@pku.edu.cn).}
\thanks{© 2022 IEEE. Personal use of this material is permitted. Permission from IEEE must be obtained for all other uses, including reprinting/republishing this material for advertising or promotional purposes, collecting new collected works for resale or redistribution to servers or lists, or reuse of any copyrighted component of this work in other works.}
}

\maketitle

\begin{abstract}
Unsupervised domain adaptation (UDA) is an approach to minimizing domain gap. Generative methods are common approaches to minimizing the domain gap of aerial images which improves the performance of the downstream tasks, e.g., cross-domain semantic segmentation. For aerial images, the digital surface model (DSM) is usually available in both the source domain and the target domain. Depth information in DSM brings external information to generative models. However, little research utilizes it. In this paper, depth-assisted ResiDualGAN (DRDG) is proposed where depth supervised loss (DSL), and depth cycle consistency loss (DCCL) are used to bring depth information into the generative model. Experimental results show that DRDG reaches state-of-the-art accuracy between generative methods in cross-domain semantic segmentation tasks. Source code is available at (https://github.com/miemieyanga/ResiDualGAN-DRDG). 
\end{abstract}


%
\IEEEpeerreviewmaketitle

\section{Introduction}

%
%
%
%
\IEEEPARstart{S}{emantic} segmentation is a classification task that gives a category to every pixel of an image. A pixel-level annotated dataset is imperative for training a deep learning semantic segmentation model. However, the pixel-level annotation of images is laborious, time-wasting, and expensive\cite{li_learning_2021,ji_generative_2021}. Most of the datasets used in the practical operation are unannotated which is impossible for training a deep learning model. A common idea is to train a model from annotated datasets and then apply it to the unannotated dataset. Nevertheless, because of the domain gap resulting from distinct data distribution between different datasets, the performance of the model trained in an annotated dataset will greatly decline when transferred to another dataset\cite{hoffman_cycada_2018}. For remote sensing (RS) images, the problem of domain gap is significantly magnified because of various imaging sensors, imaging time, weather conditions, image resolution, and so on\cite{li_learning_2021,zhao2022residualgan}. 

\begin{figure}[]
    \centering
    \includegraphics[width=.45\textwidth]{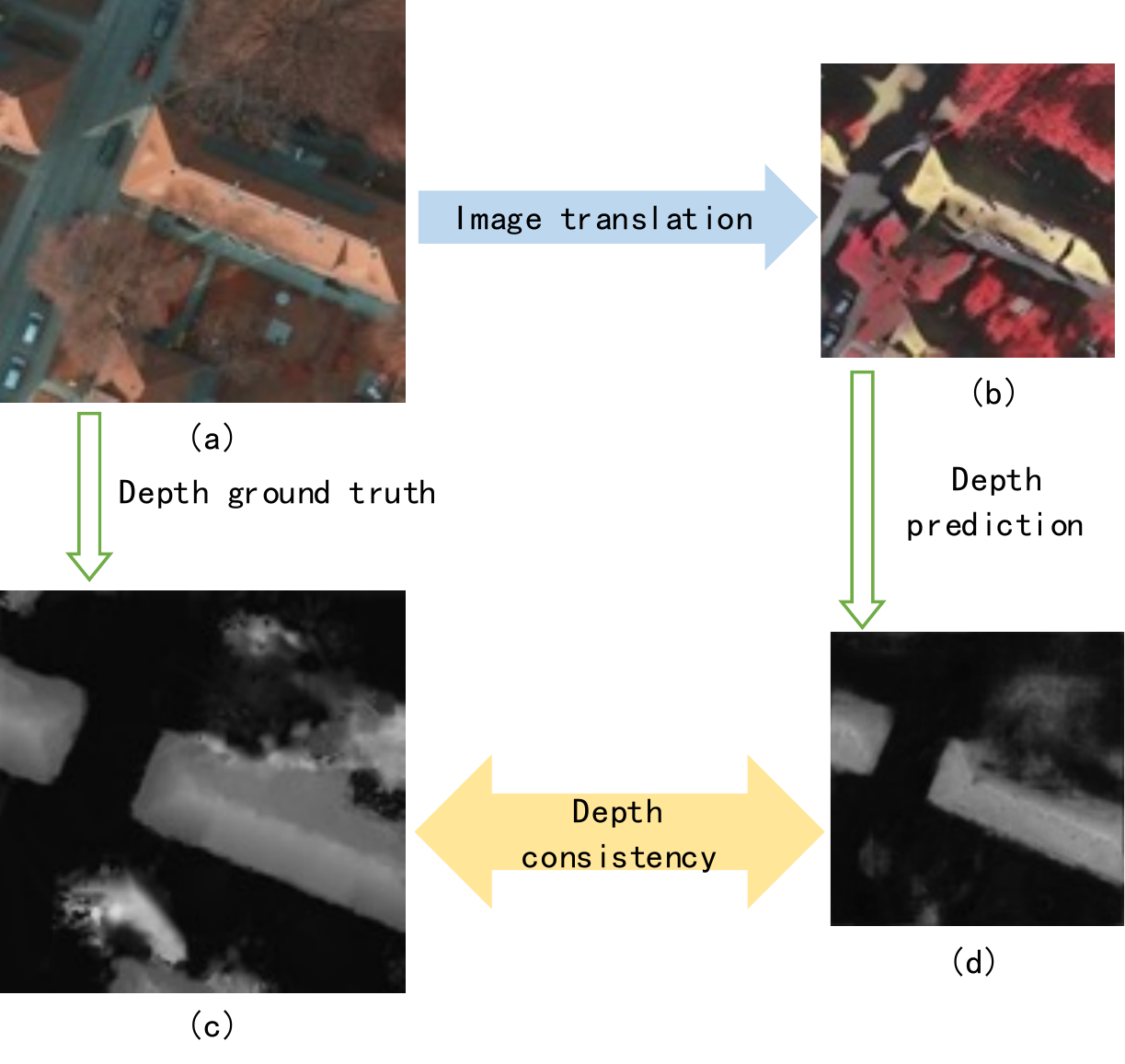}
    \caption{Depth consistency during image translation. }
    \label{fig:intro}
\end{figure}

Unsupervised domain adaptation (UDA) provides an approach for minimizing the impact of the domain gap, making it possible to transfer a deep learning model to an unannotated dataset. Methods of UDA for semantic segmentation can be roughly divided into several categories: adversarial generative methods\cite{hoffman_cycada_2018}, adversarial discriminative methods\cite{tsai_learning_2018}, and self-training methods\cite{zou_unsupervised_2018}. For RS images, adversarial generative methods, which are aimed to generate new annotated images based on generative adversarial networks (GANs), show significant advantages compared with other methods\cite{ji_generative_2021, zhao2022residualgan,gao2022cycle}. ResiDualGAN (RDG)\cite{zhao2022residualgan} is an adversarial generative method for UDA of RS images. Focusing on the unique feature of scale discrepancy and real-to-real image translation of RS images, RDG reaches state-of-the-art performance in cross-domain semantic segmentation. In this paper, RDG is selected as the foundation of our model. 

Depth information is utilized for improving the effect of UDA in some recent computer vision research. The motivation is to utilize the depth information as an external assistant which gives the deep learning model other information to improve the performance. Several works have been done by combining deep information with the generative method\cite{lee_spigan_2019}, adversarial discriminative method\cite{vu_dada_2019,saha_learning_2021}, and self-training method\cite{wang_domain_2021}, which demonstrate the effectiveness of depth information in UDA. Though improvements have been made, the defect of these methods is still nonnegligible. Depth information is usually inaccessible for the street view datasets used in these methods. An additional depth estimation model is used to give an approximate depth estimation for the street view image\cite{wang_domain_2021}, which is both inaccurate and computation-wasting. 

The digital surface model (DSM) is an elevation model which reflects the depth information of the ground surface. DSM can be easily obtained when producing an aerial image dataset by the photogrammetric method. As a result, utilizing DSM as assisted information for UDA of aerial images is reasonable. However, little research has been made on utilizing DSM to assist the UDA procedure of aerial images.

In this work, we adapt the depth information to the UDA task of semantic segmentation of aerial images. Depth-assisted ResiDualGAN (DRDG) is proposed by combining DSM with RDG where the depth supervised loss (DSL) and depth cycle consistency loss (DCCL) are used to force RDG to learn external information. The experimental results show that DRDG reaches state-of-the-art performance in open datasets. Our contribution can be summarized as follows:
\begin{itemize}
    \item A new GAN-based model, DRDG, is proposed for the UDA task of semantic segmentation of aerial images. Experimental results show that DRDG is the state-of-the-art adversarial generative UDA method for aerial images in open datasets. 
    \item DSM, which can be easily accessed for aerial images, is introduced to the DRDG as assisted information to improve the performance of the generative model, which illustrates the effectiveness of DSM for UDA of aerial images.
\end{itemize}

\section{Methodology}

\begin{figure*}[]
    \centering
    \includegraphics[width=.95\textwidth, height=.45\textheight]{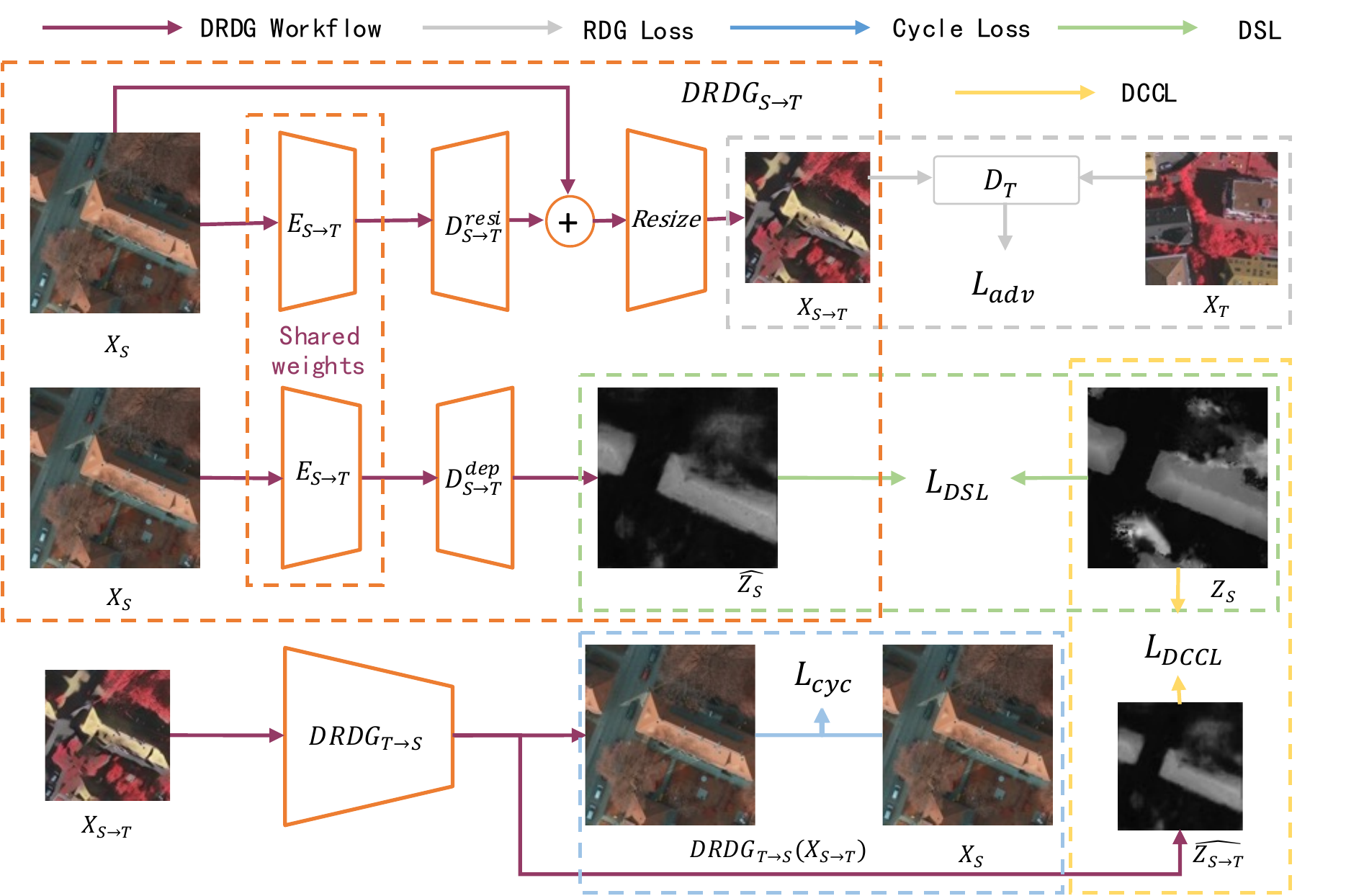}
    \caption{Overview of DRDG}
    \label{fig:drdg}
\end{figure*}

Consider a source domain $S$ and a target domain $T$, where only images in the $S$ are annotated while DSM can be obtained in both $S$ and $T$. Formally, we define $X_S\in\mathbb{R}^{H_S\times W_S\times 3}$ is the image set of $S$, $Y_S\in\{1,...,C\}^{H_S\times W_S}$ is the corresponding semantic segmentation label set where C is the number of categories, $Z_S\in\{\mathbb{Z}_{min}, \mathbb{Z}_{max}\}^{H_S\times W_S\times 1}$ is the DSM image set. Similarly, the image set in $T$ can be defined as $X_T\in\mathbb{R}^{H_T\times W_T\times 3}$. $Z_T\in\{\mathbb{Z}_{min}, \mathbb{Z}_{max}\}^{H_T\times W_T\times 1}$ is the DSM image set of $T$. Labels are inaccessible in $T$.


The objective of the proposed method is that, given the annotated $S$ and unannotated $T$, train a model $f_T$ that performs semantic segmentation task on $T$. To this end, DRDG is designed to generate target domain stylized images $x_{S\rightarrow T}\in\mathbb{R}^{H_T\times W_T\times 3}$ from source domain image $x_S$ that is annotated (Fig. \ref{fig:drdg}). In other words, DRDG minimizes the pixel-level domain gap between the source domain and the target domain. Specifically, depth information is involved in DRDG by combining two constraints, DSL and DCCL, to further improve the performance of RDG. In this section, we will introduce RDG firstly and then describe the DSL and DCCL. Finally, a semantic segmentation model $f_T$ can be trained on the annotated target domain stylized image $x_{S\rightarrow T}$. 

\subsection{ResiDualGAN}
ResiDualGAN (RDG) is an adversarial generative method for UDA of RS images, which shows prominent performance on cross-domain RS image semantic segmentation tasks. The function of RDG is to perform image-to-image translation that, in our case, transfers $x_S$ to the style of $x_T$ which minimizes the discrepancy between $S$ and $T$. Given $x_S$, $x_{S\rightarrow T}$ can be obtained by the generator of RDG, $ResiG_{S\!\rightarrow\!T}$. The procedure can be written as follows. 

\begin{equation}
\begin{split}
        x_{S\!\rightarrow\!T}=&ResiG_{S\!\rightarrow\!T}\left(x_S\right)\\
        =&Resize_{S\!\rightarrow\!T}(E_{S\!\rightarrow\!T}(D_{S\!\rightarrow\!T}^{resi}(x_S))+x_S)\
    \label{formu:rdg}
\end{split}
\end{equation}
where $Resize_{S\!\rightarrow\!T}:\mathbb{R}^{H_S{\times}W_S{\times}3}\!\rightarrow\!\mathbb{R}^{H_T{\times}W_T{\times}3}$ is a resize function that resizes images of the source domain to the size of the target domain, implemented as an bilinear interpolation in this paper, $E_{S\!\rightarrow\!T}$ is the encoder part of $ResiG_{S\!\rightarrow\!T}$ and $D_{S\!\rightarrow\!T}^{resi}$ is the decoder. 

$x_{S\rightarrow T}$ is then passed through a discriminator $D_T$, which is designed to try to distinguish whether images are generated by the generator or the initial images of the target domain. On the contrary, the objective of the generator is to generate target-stylized images to try to fool the discriminator. The opposite optimization direction between the generator and discriminator brings about an adversarial loss $L_{adv}(S, T)$. In RDG, the adversarial loss can be written as follows. 
\begin{equation}
\begin{split}
    L_{adv}&(S, T)= \\
    &\mathbb{E}_{x_T\!{\sim}\!X_T}(D_T(x_T))-
    \mathbb{E}_{x_S\!{\sim}\!X_S}(D_T({ResiG}_{S\!\rightarrow\!T}(x_S)))
    \label{formu:adv}
\end{split}    
\end{equation}

$ResiG_{T\!\rightarrow\!S}$ and $D_S$ is the inversion of the procedure above where another generator $ResiG_{T\!\rightarrow\!S}$ is implemented to reconstruct $x_{S\rightarrow T}$ to the $x_S$, and $D_S$ is used to differentiate whether images are generated by $ResiG_{T\!\rightarrow\!S}$ or original images in the source domain. The reconstruction of $x_S$ will bring into a reconstruction loss $L_{cyc}$ represented as follows. 
\begin{equation}
\begin{split}
    L_{cyc}&(S, T)=\\
    &\mathbb{E}_{x_S\!{\sim}\!X_S\!}(\Vert ResiG_{T\!\rightarrow\!S}(ResiG_{S\!\rightarrow\!T}(x_S)))-x_S \Vert_1)
    \label{formu:cyc}
\end{split}  
\end{equation}
where $\Vert \cdot \Vert_1$ is the L1 normalization. 

\subsection{Depth-Assisted ResiDualGAN}
Based on RDG, in this paper, depth supervised loss (DSL) and depth cycle consistency loss (DCCS) is involved in the generators, $DRDG_{S\!\rightarrow\!T}$ and $DRDG_{T\!\rightarrow\!S}$, to constrain the image translation procedure. The function of $DRDG_{S\!\rightarrow\!T}$ is similar to $ResiG_{S\!\rightarrow\!T}$, which is designed to perform image translation between domains. The difference is that DSM is utilized as assisted information to improve the depth correctness of image translation. 
\subsubsection{Depth Supervised Loss}
DSL is a regression loss of DSM. The DSL network shares the encoder $E_{S\!\rightarrow\!T}$ with RDG generater $ResiG_{S\!\rightarrow\!T}$ and utilizes another decoder $D_{S\!\rightarrow\!T}^{dep}$ to generate a DSM prediction $\widehat{z}_S=E_{S\!\rightarrow\!T}(D^{dep}_{S\!\rightarrow\!T}(x_S))$. Following \cite{wang_domain_2021}, a Berhu loss, which is a reverse version of Huber loss, is utilized as the measure between prediction $\widehat{z}_S$ and real DSM $z_S$:

\begin{equation}
 L_{DSL}(S) = \mathbb{E}_{(z_S,\widehat{z}_S)\!{\sim}\!(Z_S,\widehat{Z}_S)\!}(Berhu(z_S,\widehat{z}_S))
    \label{formu:dsl}
\end{equation}

where $\widehat{Z}_S$ is the predicted depth image set of $S$. The Berhu loss is:

\begin{equation}
 Berhu(z_S,\widehat{z}_S)= \left\{
    \begin{aligned}
     &|\widehat{z}_S - z_S| , &  |\widehat{z}_S - z_S|\leq L, \\
     &\frac{(\widehat{z}_S - z_S)^2 + L^2}{2L} , & |\widehat{z}_S - z_S|>L.
    \end{aligned}
    \right.
    \label{formu:berhu}
\end{equation}
where $L=0.2max(|\widehat{z}_S - z_S|)$. 

\subsubsection{Depth Cycle Consistency Loss}
In the process of image translation using RDG without depth information, only a cycle loss is utilized to constrain the pixel level cycle consistency between $S$ and $T$. As a result, images may be incorrectly translated because of the single constraints. For example, the roof may be translated as grass from $S$ to $T$, and then the incorrectly translated grass will be incorrectly translated to the roof from $T$ to $S$, where the cycle loss is minimized but the translation is bidirectionally wrong. To mitigate the bidirectionally incorrect problem, DCCS is utilized to constrain the depth cycle consistency in the image translation process. Specifically, for target stylized images $x_{S\!\rightarrow\!T}$, the corresponding depth prediction $\widehat{z}_{S\!\rightarrow\!T}$ can be generated by $DRDG_{T\!\rightarrow\!S}$. As well as the DSL, a supervised depth loss is applied to $\widehat{z}_{S\!\rightarrow\!T}$ and $z_S$, which forces the depth information immutable during the image translation process. A Berhu loss is used in DCCL:

\begin{equation}
\begin{split}
 L_{DCCL}(S,T) = \mathbb{E}_{(z_S,\widehat{z}_{S\!\rightarrow\!T})\!{\sim}\!(Z_S,\widehat{Z}_{S\!\rightarrow\!T})\!}(Berhu(z_S,\widehat{z}_{S\!\rightarrow\!T}))
    \label{formu:dccl}
\end{split}
\end{equation}

\subsubsection{Total Loss}
In the end, the total loss of image translation process can be written as:
\begin{equation}
\begin{split}
    L_{total} &= \lambda_{adv}(L_{adv}(S,T)+L_{adv}(T,S)) \\
    &+ \lambda_{cyc}(L_{cyc}(S,T)+L_{cyc}(T,S)) \\
    &+ \lambda_{DSL}(L_{DSL}(S)+L_{DSL}(T))\\
    &+ \lambda_{DCCL}(L_{DCCL}(S,T)+L_{DCCL}(T,S))\\
    \label{formu:total}
\end{split}    
\end{equation}
where $\lambda_{adv}$, $\lambda_{cyc}$, $\lambda_{DSL}$ and $\lambda_{DCCL}$ are hyper-parameters. By minimizing $L_{total}$, a well-constrained $DRDG_{S\!\rightarrow\!T}$ can be obtained to perform the image translation process to generate $X_{S\!\rightarrow\!T}$. 

After the image translation is finished, a cross-entropy loss is used to train the semantic segmentation model $f_T$:

\begin{equation}
\begin{split}
    L_{seg}=\mathbb{E}_{(x_{S\!\rightarrow\!T},y_S){\sim}(X_{S\!\rightarrow\!T},Y_S)}(CE(f_T,x_{S\!\rightarrow\!T},y_S))
    \label{formu:seg}
\end{split}    
\end{equation}
where cross-entropy loss $CE$ is:
\begin{equation}
    CE(f, x, y)=-\sum_{c=1}^C y\cdot log(f(x))
    \label{formu:fl}
\end{equation}

\subsection{Networks Settings}

\begin{table*}
\centering
\renewcommand\arraystretch{1.3}
\caption{The quantitative results of the cross-domain semantic segmentation from PotsdamIRRG to Vaihingen.}
\Huge
\label{tab:PotsdamIRRG2Vaihingen}
\resizebox{\textwidth}{!}{%
\begin{tabular}{@{}ccccccccccccccc@{}}
\toprule[5pt]
Methods &
  \multicolumn{2}{c}{\begin{tabular}[c]{@{}c@{}}Background/\\ Clutter\end{tabular}} &
  \multicolumn{2}{c}{\begin{tabular}[c]{@{}c@{}}Impervious \\ Surface\end{tabular}} &
  \multicolumn{2}{c}{Car} &
  \multicolumn{2}{c}{Tree} &
  \multicolumn{2}{c}{\begin{tabular}[c]{@{}c@{}}Low \\ Vegetation\end{tabular}} &
  \multicolumn{2}{c}{Building} &
  \multicolumn{2}{c}{Overall} \\ \toprule[5pt]
                           & IoU   & F1\_Score & IoU   & F1\_Score & IoU   & F1\_Score & IoU   & F1\_Score & IoU   & F1\_Score & IoU   & F1\_Score & mIoU   & F1\_Score \\\midrule[4pt]
Baseline                   & 2.65  & 4.41      & 43.82 & 60.56     & 12.99 & 22.47     & 49.89 & 66.35     & 24.92 & 39.59     & 57.10 & 72.38     & 31.90 & 44.29     \\ \midrule
Benjdira's\cite{benjdira_unsupervised_2019}                   & 6.93  & 9.95      & 57.41 & 72.67     & 20.74 & 33.46     & 44.31 & 61.08     & 35.60 & 52.17     & 65.71 & 79.12     & 38.45 & 51.41     \\
AdaptSegNet\cite{tsai_learning_2018}                & 5.84  & 9.01      & 62.81 & 76.88     & 29.43 & 44.83     & 55.84 & 71.45     & 40.16 & 56.87     & 70.64 & 82.66     & 44.12 & 56.95     \\
MUCSS\cite{li_learning_2021}                      & 10.82 & 14.35     & 65.81 & 79.03     & 26.19 & 40.67     & 50.60 & 66.88     & 39.73 & 56.39     & 69.16 & 81.58     & 43.72 & 56.48     \\ 
RDG\cite{zhao2022residualgan}                    & \textbf{11.22} & \textbf{18.15} & 66.93 & 80.16 & 49.46 & 66.09 & 61.74 & 76.32 & \textbf{46.64} & \textbf{63.47} & 74.52 & 85.38 & 51.75 & 64.93   \\
\midrule
DRDG (ours) & 8.49 & 14.23 & \textbf{68.74} & \textbf{81.45} & \textbf{57.54} & \textbf{72.97} & \textbf{62.91} & \textbf{77.21} & 44.22 & 61.22 & \textbf{77.67} & \textbf{87.41} & \textbf{53.26} & \textbf{65.75}    \\
  \bottomrule[5pt]
\end{tabular}%
}
\end{table*}

\begin{figure*}[ht]
    \centering
    \setlength{\abovecaptionskip}{0pt}
    \setlength{\belowcaptionskip}{0pt}
    \includegraphics[width=1.0\textwidth, height=.3\textheight]{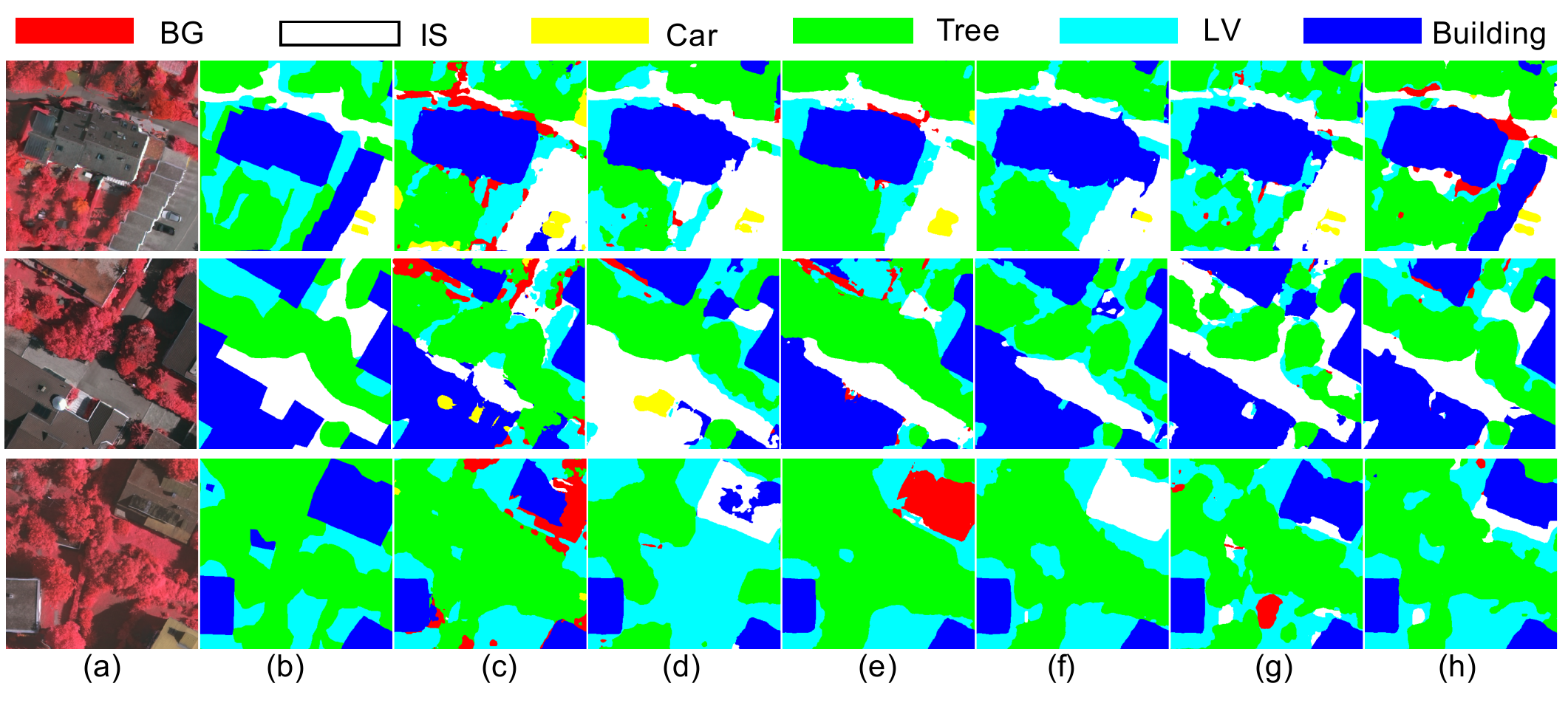}
    \caption{The qualitative results of the cross-domain semantic segmentation from PotsdamIRRG to Vaihingen. BG=Background/Cluster, IS=Impervious Surface, LV=Low Vegetation. (a) Target images. (b) Labels. (c) Baseline (DeepLabV3+). (d) CycleGAN\cite{zhu_unpaired_2017}. (e) AdaptSegNet\cite{tsai_learning_2018}. (f) MUCSS\cite{li_learning_2021}. (g) ResiDualGAN\cite{zhao2022residualgan}. (h)DRDG(ours). }
    \label{fig:PotsdamIRRG2Vaihingen}
\end{figure*}

The architecture of DRDG follows the settings of RDG\cite{zhao2022residualgan}, which is a U-shape full convolution network. The convolution kernel size is 4, the stride is 2 and padding is 1. Channels of convolution layers are set as \{64, 128, 256, 512, 512, 512, 512\}. The discrepancy between DRDG and RDG is the depth decoder in DRDG. The depth decoder is identical to the RDG decoder except for the channel of the last layer is set as 1, followed by a softmax activation to predict a depth image. The discriminator is a full convolution network with the same kernel with DRDG and channels of \{64, 128, 256, 512, 512, 1\}. DeepLabV3 is selected as our baseline of semantic segmentation. Based on thorough experiments, the hyper-parameters $(\lambda_{adv},\lambda_{cyc},\lambda_{DSL},\lambda_{DCCL})$ are set as 
$(5,10,2,1)$ respectively.

\section{Experiment Results and Discussion}

\subsection{Datasets}
Two open-source RS datasets: Potsdam and Vaihingen are used for validating the proposed method\cite{potsdam_vaihingen}. Potsdam is set as the source domain while Vaihingen is set as the target domain in the following experiments. Both of the images are produced into true orthophotos (TOPs), with annotations for 6 ground classes: clutter/background, impervious surfaces, car, tree, low vegetation, and building.  We select IR-R-G bands mode of Potsdam as PotsdamIRRG, which consist of 38 TOPs in which every TOP contains $6000\times6000$ pixels, with a resolution of 5cm. We select IR-R-G bads mode of Vaihingen which consists of 33 TOPs and every TOP contains $2000\times2000$ pixels, with a resolution of 9cm. To compare the proposed method with other methods more conveniently, we exploit the 2nd, 5th, 7th, 8th, 13th, 20th, 22nd, and 24th TOPs in Vaihingen as the validation dataset while others for test dataset, which follows other works settings. Following settings of ResiDualGAN, we also clip images of PotsdamIRRG into the size of $896\times896$ and images of Vaihingen into the size of $512\times512$. 

DSMs of PotsdamIRRG and Vaihingen are provided in the raw dataset. We follow the above processes and normalize values of DSMs into the range of 0-1. 
\subsection{Experimental Results}
To facilitate comparing with different methods, IoU and F1\_Score are employed as metrics in this paper. For every class in 6 different ground classes, the formulation of IoU can be written as:
\begin{equation}
    IoU=\frac{\left|A\cap B\right|}{\left| A\cup B \right|}
    \label{formu:11}
\end{equation}
Where A is the ground truth while B is predictions. After calculations of IoU for 6 classes, mIoU can be obtained which is the mean of IoU for every class. And the F1\_Score can be written as:
\begin{equation}
F1\_Score=\frac{2\times Precision\times Recall}{Precision+Recall}
\label{formu:12}
\end{equation}

Table \ref{tab:PotsdamIRRG2Vaihingen} shows the quantitative results of the proposed DRDG. The final result is the average of three results at the different random seeds. Four recent methods are selected for comparison, i.e., Benjdira’s, AdaptSegNet, MUCSS, and RDG. Quantitative results show that our method DRDG reaches the SOTA results between all methods by the mIou of 53.26 and F1 of 65.75 in the experiment of PotsdamIRRG to Vaihingen, which surpasses RDG by 2.9\% and 1.3\% respectively. Significantly, DRDG reaches the mIou of 77.67 and F1 of 87.41 in the building class, which surpasses RDG by 4.2\% and 2.4\% respectively. As shown in Fig. \ref{fig:intro}, the outline of buildings is distinguishable, which may contribute to improving the accuracy of the building class through the constrain of DSL and DCCL. 
Fig. \ref{fig:PotsdamIRRG2Vaihingen} is the qualitative results of DRDG, which also shows the superiority of our method. 

\subsection{Ablation Study}
Table \ref{tab:ablation} shows the ablation study of DRDG. We performed the experiments of removing DCCL (w/o DCCL), removing DSL (w/o DSL), and removing both DCCL and DSL (RDG). Experimental results show that both DCCL and DSL contribute to the improvements of DRDG. Significantly, if only DCCL is involved, the mIoU and F1 decline to 50.33 and 63.39 respectively. It is plausible that DCCL is designed to constrain the depth consistency but to bring depth information into the model. As a result, the $\lambda_{DCCL}$ is set smaller than $\lambda_{DSL}$ to ensure DSL brings correct depth information into the model. 
\begin{table}[]
\centering
\caption{Ablation results for DRDG. }
\renewcommand\arraystretch{1.3}
\label{tab:ablation}
\resizebox{.48\textwidth}{!}{%
\begin{tabular}{@{}lllll@{}}
\toprule[1.3pt]
               & DSL & DCCL & mIoU & F1 mean \\ \midrule[1pt]
RDG\cite{zhao2022residualgan}              &     &      & 51.75     &  64.93        \\
DRDG(w/o DSL)  &     &  \checkmark    & 50.33      & 63.39         \\
DRDG(w/o DCCL) & \checkmark    &      & 52.93       &  65.56         \\
DRDG           &  \checkmark   & \checkmark     & \textbf{53.26} & \textbf{65.75}       \\ \bottomrule[1.3pt]
\end{tabular}%
}
\end{table}
\section{Conclusion}
In this paper, a depth-assisted GAN-based framework, DRDG, is proposed for aerial image cross-domain semantic segmentation. The DSM information is introduced into the GAN process by two constraints: DSL and DCCL, which force GAN models to maintain the DSM correctness during generating images. The experimental results show the priority of the proposed method, which reaches the state-of-the-art performance of the generative method in the PotsdamIRRG to Vaihingen cross-domain segmentation task.

There is still a lot of work to be done in our paper. The first is to resolve the instability of the GAN-based method. Although RDG has made a progress on stability, however, the results still fluctuate greatly at different random initialization. The second is to find an effective way to directly combine the DSM with the segmentation model. Training the model through an end-to-end approach instead of a multi-stage framework may yield a more stable result. The last is to utilize self-training methods to further improve the performance. DSMs are accessible in both the source domain and the target domain, which facilitates self-training strategies to initialize the weights of models.  

\section*{Acknowledgment}
Thanks for the support from Professor Ge Li at Peking University and computation resources from VirtAI Tech.

\ifCLASSOPTIONcaptionsoff
  \newpage
\fi



\bibliographystyle{IEEEtran}
\bibliography{./uda-pi.bib, ./residualgan.bib}
\end{document}